\documentclass{article}

\usepackage{arxiv}

\usepackage[utf8]{inputenc} 
\usepackage[T1]{fontenc}    
\usepackage{hyperref}       
\usepackage{url}            
\usepackage{booktabs}       
\usepackage{amsfonts}       
\usepackage{nicefrac}       
\usepackage{microtype}      
\usepackage{lipsum}		
\usepackage{amsmath,amsfonts}
\usepackage{algorithmic}
\usepackage{algorithm}
\usepackage{array}
\usepackage[caption=false,font=normalsize,labelfont=sf,textfont=sf]{subfig}
\usepackage{textcomp}
\usepackage{stfloats}
\usepackage{verbatim}
\usepackage{multirow} 
\usepackage{graphicx}
\usepackage{booktabs}
\usepackage{array}
\usepackage{cite}
\usepackage{hhline}
\usepackage{caption}
\usepackage{subcaption}

\title{Student Mental Health Screening via Fitbit Data Collected During the COVID-19 Pandemic}

\author{Rebecca Lopez\\
        Department of Data Science\\
        Worcester Polytechnic Institute\\
        Worcester, MA 01609 USA \\
	\texttt{rlopez2@wpi.edu} \\
	\And
	Avantika Shrestha\\
    Department of Data Science\\
    Worcester Polytechnic Institute\\
    Worcester, MA 01609 USA \\
	\texttt{ashrestha4@wpi.edu} \\
	\And
	ML Tlachac\\
    Department of Information \\ Systems and Analytics\\
    Bryant University\\
    Smithfield, RI 02911 USA \\
	\texttt{mltlachac@bryant.edu} \\
	\And
	Kevin Hickey\\
    Department of Data Science\\
    Worcester Polytechnic Institute\\
    Worcester, MA 01609 USA \\
	\texttt{khickey@wpi.edu} \\
	\And
	Xingtong Guo\\
    Department of Civil, Environmental,\\ and Architectural Engineering\\
    Worcester Polytechnic Institute\\
    Worcester, MA 01609 USA \\
	\texttt{xguo3@wpi.edu} \\
	\And
	Shichao Liu\\
    Department of Civil, Environmental,\\ and Architectural Engineering\\
    Worcester Polytechnic Institute\\
    Worcester, MA 01609 USA \\
	\texttt{sliu8@wpi.edu} \\
	\And
	Elke Rundensteiner\\
    Department of Data Science\\
    Department of Computer Science\\
    Worcester Polytechnic Institute\\
    Worcester, MA 01609 USA \\
	\texttt{rundenst@wpi.edu} \\
}

\date{}

\renewcommand{\shorttitle}{Student Mental Health Screening via Fitbit}

\hypersetup{
pdftitle={ },
pdfsubject={ },
pdfauthor={ },
pdfkeywords={ },
}

\begin{document}
\maketitle

\begin{abstract}
College students experience many stressors, resulting in high levels of anxiety and depression. Wearable technology provides unobtrusive sensor data that can be used for the early detection of mental illness. However, current research is limited concerning the variety of psychological instruments administered, physiological modalities, and time series parameters. In this research, we collect the Student Mental and Environmental Health (StudentMEH) Fitbit dataset from students at our institution during the pandemic. We provide a comprehensive assessment of the ability of predictive machine learning models to screen for depression, anxiety, and stress using different Fitbit modalities. Our findings indicate potential in physiological modalities such as heart rate and sleep to screen for mental illness with the F1 scores as high as 0.79 for anxiety, the former modality reaching 0.77 for stress screening, and the latter modality achieving 0.78 for depression. This research highlights the potential of wearable devices to support continuous mental health monitoring, the importance of identifying best data aggregation levels and appropriate modalities for screening for different mental ailments.
\end{abstract}


\section{Introduction} \label{sec:introduction}

\subsection{Burden and Prevalence of Mental Illnesses} There is a global mental health crisis \cite{demyttenaere2006comorbid, proudman2021growing}, which was exacerbated by the COVID-19 pandemic \cite{czeisler2020mental}. Mental illnesses affect millions of individuals annually, decreasing their quality of life and contributing to substantial financial burdens \cite{kessler2008individual,demyttenaere2006comorbid}. In addition to directly increasing healthcare costs, there are indirect costs such as lost wages \cite{kessler2008individual}. Mental distress affects role functioning capabilities of the individual \cite{lecrubier2001burden}, which is especially problematic for college students \cite{conley2020navigating}. Therefore, identifying and adequately addressing mental health symptoms is critical for improving individuals' quality of life. 

Major Depressive Disorder (MDD) \cite{bains2023} and State Anxiety (SA) \cite{spielberger1983, Knowles2020} are among the most prevalent mental illnesses. MDD is characterized by persistent sadness and lack of interest in previously enjoyed activities \cite{bains2023}. Meanwhile, SA refers to the emotional state of heightened apprehension related to specific situations \cite{spielberger1983, Knowles2020}. Both can lead to impairment in daily abilities and increased risk of developing more chronic conditions, thereby further exacerbating health care costs \cite{voinov2013}. 

A shared symptom of these two mental illnesses is Perceived Stress (PS) \cite{CohenPSS}. Often measured as the degree to which situations in one’s life are considered stressful, PS has been linked to significant negative health outcomes \cite{CohenPSS}. College students in particular experience high stress levels due to the many changes in their lives as they transfer into, through, and out of college \cite{cage2021student}; this stress contribute to the high rates of depression and anxiety among student populations. 

\subsection{Barriers to Diagnosis and Digital Screening} 
There is a significant shortage of qualified medical experts for mental illness assessment \cite{andrilla2018geographic}. The demand for mental health services often exceeds availability, which is particularly true for many college mental health counseling centers \cite{eab}. Access to mental health services is also limited given rising healthcare costs \cite{Coombs2021,APA2022}. Early recognition and treatment of mental illnesses can greatly reduce their health, financial, and social burdens \cite{halfin2007depression}. Unfortunately, mental illness diagnosis is an arduous \cite{Bohman2023}, and patients do not always recognize when they need to seek help \cite{epstein2010didn}. The availability, access, and recognition challenges reinforce the need for alternative solutions to improve mental health screening. 

To combat these issues, the research community \cite{dwyer2018machine,bryan2024behind,Khoo2024} has applied predictive machine learning to identify early signs of mental health issues from ubiquitous mobile and wearable devices like Fitbit smartwatches \cite{Fitbit2023}. By detecting patterns in physiological activities, these models can facilitate early detection, leading to timely and thus more likely effective interventions \cite{halfin2007depression}. Proactive interventions have the potential to reduce severity and shorten the duration of these conditions \cite{McGorry2022}. In particular, predictive models could increase access and minimize variability in assessing mental illnesses \cite{Ghosh2022}. 

Moreover, utilizing machine learning in conjunction with wearable technology could increase mental health screening in areas of high demand (such as college campuses \cite{eab}) and with low coverage of health services \cite{watkins2012increased}. Unlike traditional screening methods that require considerable human resources such as trained personnel and infrastructure, wearable technologies promise efficient scaling and can reach broader populations, including those in underdeveloped areas \cite{Seshadri2020, Chan2014}. A positive screening result could schedule the user with a clinician via telehealth \cite{koonin2020trends} and/or launch a mobile intervention \cite{baxter2020assessment,bakker2018randomized}. 

\begin{figure}[ht]
  \centering
  \includegraphics[width=0.85\columnwidth, trim=80 100 40 80, clip]{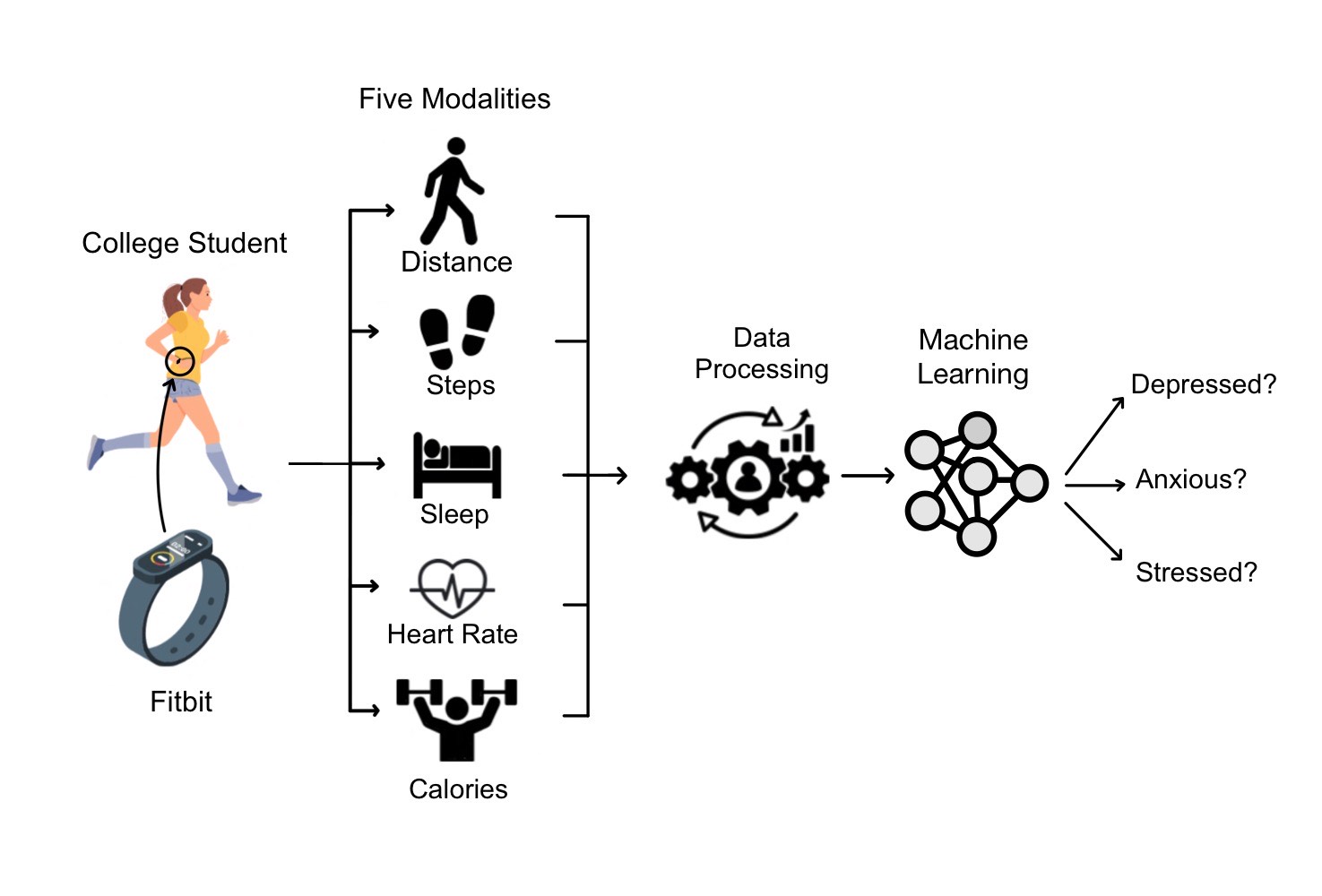}
  \caption{Screening for Mental Illnesses using Fitbit Data Modalities.}
  \label{fig:highlevel}
\end{figure}

Continuous non-intrusive collection of a user's physiological state can provide an in-depth view into their behavior and mental health \cite{torous2016new,bryan2024behind,Khoo2024}, as depicted in Fig. \ref{fig:highlevel}. Thus, machine learning offers the unique opportunity for a personalized approach by analyzing how mental illness symptoms manifest through physiological data, enhancing the effectiveness of the treatment strategies \cite{Annabestani2024}. 
 
\subsection{Our Approach and Contributions}
To explore the relationships between mental distress signals and physiological responses, we collected the StudentMEH (Student Mental and Environmental Health) dataset from an undergraduate college student population during the early stages of the COVID-19 pandemic \cite{Cucinotta2020}. The data collected  includes  measures related to student well-being,  indoor environmental conditions, environmental satisfaction, and physiological factors such as heart rate and step count from Fitbit devices. We focus on college students due to their high rates of depression, anxiety, and stress \cite{cage2021student,eab}. 

We then aim to investigate how physiological data patterns may aid in screening college students for MDD, SA, and/or PS. As such, 
a primary goal of this research is to identify the most effective Fitbit data modalities for  MDD, SA, and PS screening models. Furthermore, we explore the optimal aggregation level of time-series data from wearable devices, a topic that has not been thoroughly investigated for MDD, SA, and PS. Studies suggest that anxiety and stress levels can fluctuate within a single day \cite{Othman2019, Geschwind2010}, making it essential to map these patterns using finer time granularities. Determining the optimal data types and the necessary granularity of data aggregation enables more precise and efficient diagnostic tools. 

As our preliminary analysis \cite{bigdata} only screened for stress with a single type of classifier, we now expand to include multiple mental health conditions and classifiers. As such, our detailed comparative analysis not only enhances the understanding of how different data inputs affect model performance but also informs the selection of best techniques for clinical applications, thus enabling the optimization of scarce mental health resources.The contributions of our research include:

\begin{enumerate}
\item Collection of the StudentMEH Fitbit dataset containing sensor data and mental health survey data from college students during the COVID-19 pandemic.
\item Classification of five physiological modalities for three different mental illness labels: MDD, SA, and PS. 
\item Analysis of impact of data aggregation at numerous alternate time granularities on mental health screening. 
\item Comparative evaluation of popular predictive machine learning models across modalities and time granularities.  
\end{enumerate}

\section{Related Literature}

Prior studies have explored the effectiveness of leveraging smartphone and wearable technology sensor data for mental health symptoms and disorder assessment \cite{abd2023systematic,bryan2024behind,Khoo2024,melcher2020digital,sevil2021discrimination}. Initially, smartphones were used for such research due to their earlier release and subsequent adoption \cite{bryan2024behind}. 
Madan et al. \cite{madan2011sensing} extracted location and communication logs from student smartphones to assess self-reported feelings of stress and depression. Since then, research has leveraged a variety of smartphone data including location \cite{boukhechba2018demonicsalmon,wang2014studentlife,nepal2022covid,stamatis2024differential}, communication logs \cite{tlachac2022deprest,tlachac2024symptom,boukhechba2018demonicsalmon,WARE2022100356,stamatis2024differential,zhao2024bayesian}, and voice recordings \cite{tlachac2022studentsadd,tasnim2024machine,zhao2024bayesian} to assess depression \cite{tasnim2024machine,tlachac2022studentsadd,boukhechba2018demonicsalmon,wang2014studentlife,tlachac2022deprest,tlachac2024symptom,stamatis2024differential,nepal2022covid,WARE2022100356}, anxiety \cite{tasnim2024machine,boukhechba2018demonicsalmon,tlachac2022deprest,tlachac2024symptom,stamatis2024differential,nepal2022covid}, and stress \cite{tasnim2024machine,wang2014studentlife,nepal2022covid}. Many studies notably include student populations \cite{boukhechba2018demonicsalmon,wang2014studentlife,tlachac2022studentsadd,tasnim2024machine,nepal2022covid,WARE2022100356}. 

Research with wearable devices has been increasingly featured in the related literature on mental health assessment, largely 
due to the ability of wearable technology to continuously monitor physiological data.  Both smartphones and wearable devices can collect  various types of sensor data from acceleration to light exposure \cite{Khoo2024,abd2023systematic}, while sleep, heart rate, and circadian rhythms are more easily collected by wearable devices \cite{abd2023systematic}. Modalities including exercise, steps, and calories can then be extracted from wearable technology sensor data.  Smartwatches in particular have become a popular method for data collection with 33 studies identified that used  Actiwatch or Fitbit between 2015 and 2022 \cite{abd2023systematic}.

There are a number of studies that collected sensor data from both smartphones and smartwatches simultaneously for mental health assessment. Among them, Lu et al. \cite{lu2018joint} used predictive modeling to assess depression in college students. Meanwhile, Wang et al. \cite{wang2018tracking} and Xu et al. \cite{xu2023globem} both found that sleep and mobility patterns correlate with depression screening scores for college students. Likewise, Sano et al. \cite{Sano2018} found sleep and mobility patterns to be predictive of student stress. Unlike the other studies, Moshe et al. \cite{Moshe2021} collected depression, anxiety, and stress labels for adults aged 24-80. They found associations between sleep, depression, and anxiety as well as between heart rate variability and anxiety \cite{Moshe2021}. Zhang et al. \cite{zhang2021relationship} also analyzed sensor data from adults, but focused solely on stress prediction. 

Fewer studies modeled only wearable technology sensors for affective computing. For example, Rykov et al. \cite{rykov2021digital} leveraged Fitbit data labeled with depression screening scores in both explanatory and predictive models. Further, Dai et al. \cite{Dai2023} utilized Fitbit data from the  ``All of Us'' dataset to screen for mental illness. However, depression and anxiety were grouped into one common label instead of distinguishing the type of mental illness in the predictions.
 
Prior research has shown that anxiety, depression, and stress in university students can vary based on daily activities and social interactions which highlights the importance of monitoring more frequently to capture the dynamic nature of these mental health issues \cite{Othman2019, Geschwind2010}. While the most useful time granularity of communication logs for effective depression screening has been explored \cite{tlachac2021mobile}, the most effective time granularity of Fitbit data remains unexplored. Further, despite MDD being a focal point in related research \cite{bryan2024behind}, gaps persist in the research on stress and anxiety. Our research addresses these knowledge gaps by exploring alternate time granularities and three mental health conditions with physiological modalities. 

\section{The Student Fitbit Dataset}

Driven by our desire to gain an understanding about the impact of the pandemic on our college students, our team has collected the Student Mental and Environmental Health (StudentMEH) \cite{liu2025studentmeh} from  166 college students aged 18-23 over a period spanning from May 15, 2020 to May 26, 2021 amidst the COVID-19 pandemic. Notably, the World Health Organization declared a global pandemic on March 11, 2020. The StudentMEH dataset contains five Fitbit sensor data modalities (sleep, distance traveled, heart rate, step, and calories burned) and self-reported assessment measures for three mental health conditions (depression, anxiety, and stress). 

To safeguard participants' privacy, we created Fitbit accounts using random unique ID numbers not linked to participants' names nor email addresses. Similarly, for survey data, participants' emails were replaced with these ID numbers, ensuring they  remain confidential during data analysis. In total, over 2 million minutes of sensor data was  collected. Out of the 166 total participants, 155 answered the demographic questions with 68.8\% identifying as `Female,' 30.5\% identifying as `Male,' and 0.7\% selecting `Prefer not to answer'. 

\subsection{Self-Reported Measures} \label{sec: measures}
Participants were periodically prompted to report on their psychological and physical health parameters including stress levels, anxiety, depression, indoor environmental quality satisfaction, and self-reported learning performance. The surveys included the Center for Epidemiologic Studies Depression Scale 10-item  (CES-D-10) \cite{Andresen1994} and  State-Trait Anxiety Inventory (STAI) \cite{Ercan}, administered monthly, and Perceived Stress Scale 4-item version (PSS-4) \cite{CohenPSS}, administered weekly using an online survey tool. While self-reported data may have bias and clinician assessment would be preferred, our labeling approach is practical for initial model development. 

\begin{figure*}[t]
    \centering

    \begin{minipage}[t]{0.32\textwidth}
        \centering
        \includegraphics[width=\linewidth]{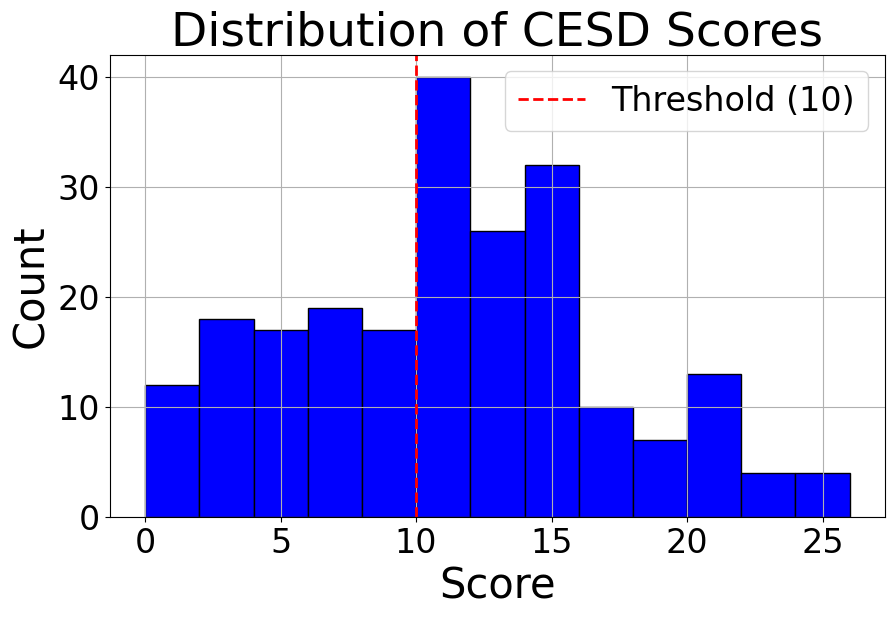}
        \subfloat{Depression Screening Scores}
        \label{fig:dist-depression}
    \end{minipage}\hfill
    \begin{minipage}[t]{0.32\textwidth}
        \centering
        \includegraphics[width=\linewidth]{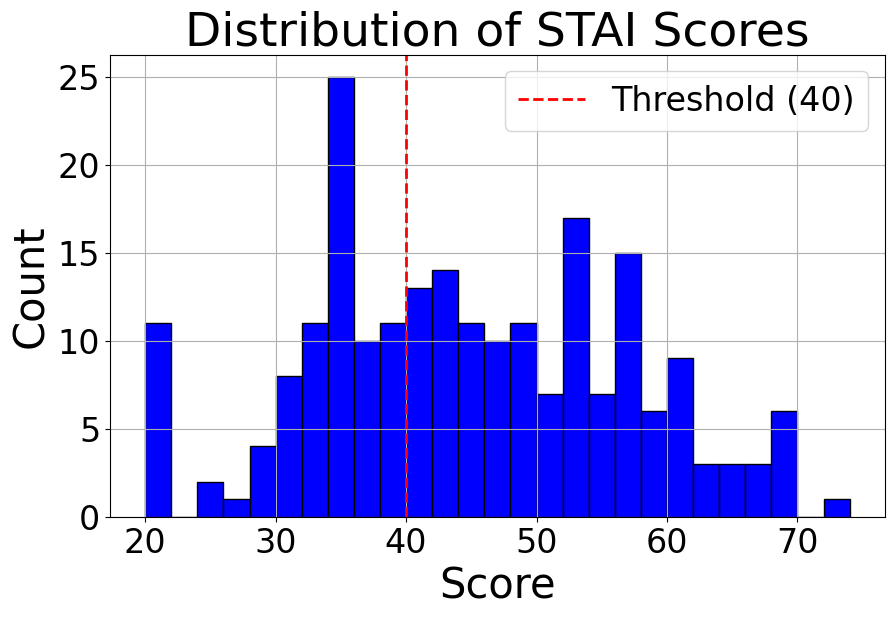}
        \subfloat{Anxiety Screening Scores}
        \label{fig:dist-anxiety}
    \end{minipage}\hfill
    \begin{minipage}[t]{0.32\textwidth}
        \centering
        \includegraphics[width=\linewidth]{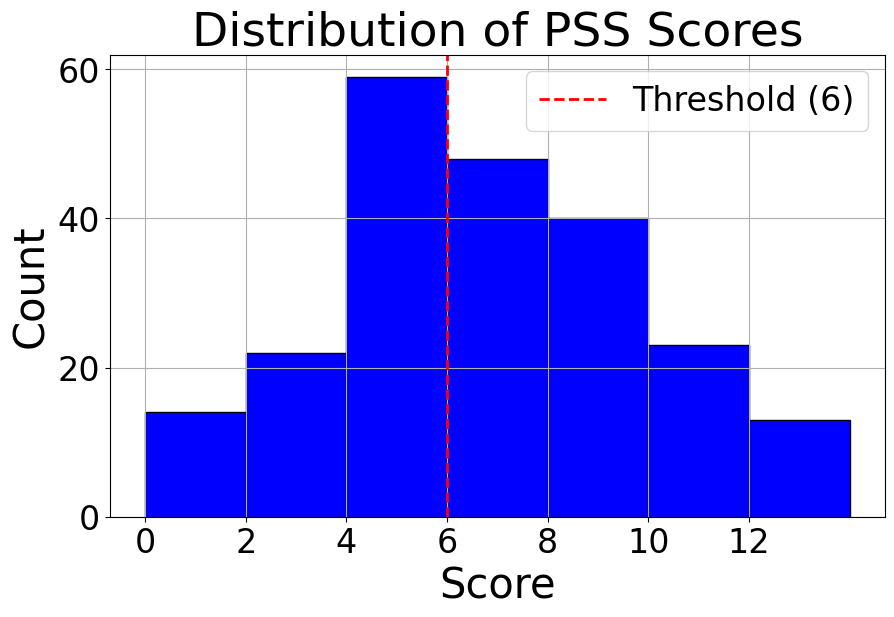}
        \subfloat{Stress Screening Scores}
        \label{fig:dist-stress}
    \end{minipage}

    \vspace{10pt}
    \caption{Distributions of CES-D, STAI, and PSS scores. Dashed line indicates the binary classification threshold.}
    \label{fig:combined}
\end{figure*}

To categorize the symptoms measured by the three psychiatric instruments, we used established cutoffs \cite{Andresen1994, Warttig2013}.  A score of at least 10 for the CES-D \cite{Andresen1994} indicates depression, a score of at least 40 for the STAI  \cite{Ercan, Grabowska2024} signifies clinical levels of anxiety, and a score of at least 6 for the PSS-4 \cite{Warttig2013} denotes a high level of perceived stress. These thresholds, 
widely utilized in psychological research, are representative of the general population \cite{Andresen1994, Warttig2013}. As displayed in Figure \ref{fig:combined}, 53.7\%, 60.6\%, and 60.8\% of student participants screened positive for depression, anxiety, and stress, respectively. 

\begin{figure*}[ht]
  \centering
  \includegraphics[width=0.96\textwidth, trim=20 70 30 80, clip]{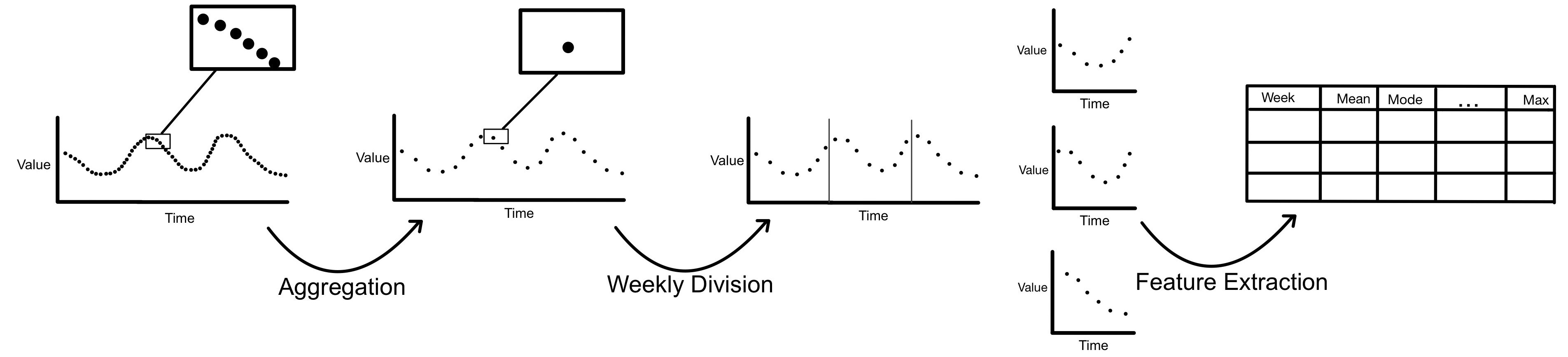}
    \caption{Data Processing Pipeline: (a) Aggregation of fine-grained irregular time-series data into a unified coarser-grained granularity of 1 or more hours, with imputation performed during aggregation and experimentation with duration 
    units  (1, 4, 6, 8, 12, 24 hour aggregation choices); (b) Partitioning into  week-long time series augmented by illness labels.
    (c) Transformation of each weekly time series into a record of derived  features.}
  \label{fig:data_processing} 
\end{figure*} 

\subsection{Wearable Sensor Data}
Participants were asked to wear Fitbit \cite{Fitbit2023} devices which continuously collected physical activity and physiological data. Data was recorded relating to calories burned, distance traveled, step count, and sleep stages on a minute-by-minute basis. Likewise, heart rate data was recorded every five seconds. These Fitbit modalities can offer insights into mental health by reflecting overall physical and emotional well-being. For instance, reduced step count and reduced distance traveled are common in individuals with depression or anxiety \cite{dang2023role, deangel2022digital}. Abnormal heart rate variability is linked to stress and mood disorders, while lower calories burned correspond to decreased activity in depression \cite{deangel2022digital}. Disrupted sleep patterns, including less deep sleep, are  prevalent in  mental illnesses  \cite{deangel2022digital}. Together, these  non-invasive metrics provide a well-rounded view of an individual's health, making them promising indicators for mental illness screening.

Fitbit devices use a combination of sensors and algorithms to produce the five modalities. The step count and distance traveled are measured using a 3-axis accelerometer \cite{Feehan2018}. In addition to using the intensity and duration of activities recorded by the accelerometer, calories burned are estimated using basal metabolic rate (BMR) calculations, which consider user's age, gender, weight, and height. Meanwhile, heart rate data is recorded using photoplethysmography (PPG) sensors \cite{Weiler2017}, which emit green light and measure the amount of light absorbed by the blood in the wrist. Sleep stage data is collected using a combination of accelerometer data and heart rate variability. Fitbit's sleep tracking algorithms analyze movement and variations in heart rate to determine sleep stages \cite{Fitbit2023}, including light sleep, deep sleep, and REM sleep.

Overall, 2,059,371 minutes of data was collected. Inclusion criteria for this analysis required participants to provide at least one week of continuous Fitbit data. Out of the initial cohort of 166 participants, 88 met this requirement.

\section{Methodology} \label{section.data.processing}

\subsection{Data Cleaning and Imputation}
For heart rate and calories burned, values of zero were treated as missing because they have a minimum threshold in practice. Missing values for all but sleep modalities were imputed using linear interpolation to ensure a logical progression in the recorded data. Linear interpolation was employed as the data is subsequently aggregated over extended periods, mitigating  minor inaccuracies introduced by interpolation. Regarding sleep data, we assume that missing values following and before an awake status indicate the participant is awake. This assumption allowed us to maintain a realistic representation of sleep patterns and activity levels.

\subsection{Aggregation Design Choices}
We employ   a data aggregation strategy to compress the raw time series data collected from wearable devices into manageable data sizes. For metrics such as calories burned, steps taken, and distance traveled, we aggregate the raw data by summing the values within each predefined aggregation period. For example, we sum these values to capture the total physical activity within a 12-hour period, i.e. day and night. This approach provides a consolidated measure of physical activity levels over specific time intervals. We recognize that summing the data may miss short-term variations. To mitigate this, we select various aggregation levels based on prior studies \cite{WARE2022100356,xu2022globem,Tlachac2021,Sano2018} that capture meaningful  trends, providing a balance between detail and comprehensibility.

For heart rate data, we calculate the mean value per time interval to represent cardiovascular activity within each segment. For sleep data, we quantify the occurrences of different sleep stages within each segment by calculating the total count of each specific stage, thereby gaining insights into sleep quality and patterns \cite{miller2022validation}. Lastly, we also compute the standard deviation for each modality. This helps in assessing the variability and stability of the data across different periods. 

\subsection{Weekly Segmentation and Feature Extraction}
We segment each participant's data into weekly intervals -- each spanning from Monday to Sunday. Each weekly segment is assigned a label derived from the results of three different assessment instruments, specifically those from the survey conducted closest to the corresponding week. An individual with multiple weeks of data will thus have several distinct data instances, each associated with the labels from the survey data nearest in time to that particular week. This  ensures that the data is temporally aligned with the most relevant survey responses allowing for an accurate analysis of the relationship between weekly data and their survey outcomes.

Rather than utilizing the aggregated time series values in each weekly segment directly (Section \ref{section.data.processing}), we compute statistical features from each time series segment, detailed in to Table \ref{table-statistical-features}. Replacing the time series by a structured feature record  provides notable advantages. It converts the variable-length time series into a fixed set of summary statistics ready for consumption by classical ML models. Additionally, extracted statistical features can be more easily interpretable \cite{guo2020evaluation}.

As preliminary  experiments using the full suite of features from the Time Series Feature Extraction Library \cite{tsfel}  yielded low-performing models and unstable results, we select a subset of classical functions from this library to compute statistical features of  weekly time series segments that led to more promising results (Refer to Table \ref{table-statistical-features}). They provide a comprehensive summary of the data's  variability, distribution shape, complexity, and central tendency. 

\begin{table}[ht]
\caption{Feature Engineering Using Statistical Functions.}
\centering
\begin{tabular}{|>{\centering\arraybackslash}m{3cm}|>{\centering\arraybackslash}m{5cm}|}
\hline
\textbf{Measure} & \textbf{Mathematical Notation} \\
\hline
Mean & $\frac{1}{n} \sum_{i=1}^n x_i$ \\
\hline
Mode & $\text{argmax}_x \, \text{count}(x)$ \\
\hline
Median & 
\begingroup
\renewcommand{\arraystretch}{0.9} 
\setlength{\arraycolsep}{2pt}    
$\left\{ \begin{array}{ll}
    x_{(n+1)/2} & n \text{ odd} \\
    \frac{x_{n/2} + x_{n/2+1}}{2} & n \text{ even}
\end{array} \right.$
\endgroup\\
\hline
Standard Deviation & $\sqrt{\frac{1}{n} \sum_{i=1}^n (x_i - \bar{x})^2}$ \\
\hline
Variance & $\frac{1}{n} \sum_{i=1}^n (x_i - \bar{x})^2$ \\
\hline
Range & $x_{\text{max}} - x_{\text{min}}$ \\
\hline
Inter Quartile Range & $Q_3 - Q_1$ \\
\hline
First Quartile (Q1) & $x_{(n+1)/4}$ \\
\hline
Third Quartile (Q3) & $x_{3(n+1)/4}$ \\
\hline
Sum & $\sum_{i=1}^n x_i$ \\
\hline
Unique Values & $|\{x_1, x_2, \ldots, x_n\}|$ \\
\hline
Minimum & $\min(x_1, x_2, \ldots, x_n)$ \\
\hline
Maximum & $\max(x_1, x_2, \ldots, x_n)$ \\
\hline
Root Mean Square  & $\sqrt{\frac{1}{n} \sum_{i=1}^n x_i^2}$ \\
\hline
Entropy & $- \sum_{i=1}^n p_i \log(p_i)$ \\
\hline
\end{tabular}
\label{table-statistical-features}
\end{table}

\subsection{Machine Learning Models}

We employ a variety of popular machine learning models that have performed well on similar sensor data  \cite{Khoo2024} to classify MDD, SA, and PS. Classical models  require less computing infrastructure -- making them easier to deploy. They also allow for more interpretability  into which features were used for decision-making. By focusing on these, we develop a baseline against which future deep learning models can be compared.

A {\it Decision Tree (DT)} splits the data set into subsets  based on  input features, creating branches until the data is adequately separated or a stopping condition is met. The uncertainty in the data is measured by entropy:
\begin{equation}
    H(Y) = -\sum_{i=1}^C p_i \log(p_i)
\end{equation}
where \(H(Y)\) is the entropy, \(C\)  the number of classes, and \(p_i\)  the probability of class \(i\). The best feature to split on is determined by the reduction in entropy:
\begin{equation}
    IG(Y, X) = H(Y) - H(Y|X)
\end{equation}
where \(IG(Y, X)\) is the information gain of feature \(X\) with respect to the target variable \(Y\), \(H(Y)\)  the entropy of \(Y\) before the split, and \(H(Y|X)\)  the conditional entropy of \(Y\) given \(X\).

{\it
Logistic Regression (LR)} predicts the probability of a binary outcome based on  predictor variables. It uses a logistic function to model the relationship between input features and outcome probability:
\begin{equation}
    P(Y=1|X) = \frac{1}{1 + e^{-(\beta_0 + \beta_1 x_1 + \cdots + \beta_p x_p)}}
\end{equation}
where \(P(Y=1|X)\) is the probability of the target variable \(Y\) being 1 given the input features \(X\), \(\beta_0\) is the intercept, and \(\beta_1, \ldots, \beta_p\) are the coefficients of the model.

{\it Random Forest (RF)} is an ensemble method that builds many decision trees during training, where the final prediction is the mode of all trees:
\begin{equation}
    \hat{Y} = \text{mode}\{h_t(X)\}_{t=1}^T
\end{equation}
where \(\hat{Y}\) is the predicted class, \(h_t(X)\) is the prediction of the \(t\)-th decision tree, and \(T\) is the total number of trees.

{\it Support Vector Machine (SVM)} finds the best boundary maximizes the distance between the boundary and the nearest data points of each class (support vectors):
\begin{equation}
    f(X) = \text{sign}\left( \sum_{i=1}^n \alpha_i y_i K(X_i, X) + b \right)
\end{equation}
where \(f(X)\) is the decision function, \(\alpha_i\) are coefficients, \(y_i\) are class labels, \(K(X_i, X)\) the kernel function that maps input features to a higher-dimensional space, and \(b\) the bias term.

{\it
XGBoost} \cite{chen2016xgboost} is a  boosting method that builds an ensemble of trees sequentially. Each new tree is trained to correct errors made by the prior trees, focusing on the residual errors:

\begin{equation}
    \hat{Y}_i = \sum_{k=1}^K f_k(X_i), \quad f_k \in \mathcal{F}
\end{equation}
where \(\hat{Y}_i\) is the predicted value for instance \(i\), \(f_k\) the \(k\)-th tree in the ensemble, \(X_i\) the input features for instance \(i\), and \(\mathcal{F}\)  the set of all possible trees.

{\it
AdaBoost} \cite{freund1997decision} is a boosting technique that combines the performance of weak classifiers to produce a strong ensemble by iteratively training classifiers, each time focusing more on previously misclassified ones.
The prediction is given by:
\begin{equation}
    \hat{Y} = \text{sign}\left( \sum_{t=1}^T \alpha_t h_t(X) \right)
\end{equation}

where \(\hat{Y}\) is the predicted class, \(\alpha_t\) the weight for classifier \(h_t\), and \(T\)  the number of classifiers.

\subsection{Feature Selection and Hyperparameter Tuning}
For the  reduction of redundant features, we use the standard recursive feature elimination strategy across all models. Initially, we fit a tree-based method using all features, subsequently eliminating features with lower importance \cite{guyon2002gene,pedregosa2011scikit}. This process of  elimination continues until a decline in model performance is observed, 
indicating a near-optimal subset of features. We employ Recursive Feature Elimination with Cross-Validation (RFECV) using scikit-learn \cite{pedregosa2011scikit}.  

Following prior research's  validated approach \cite{Adhinegoro2023, Xu2024} for optimizing  model hyperparameters, we  employ the state-of-the-art hyperparameter optimization framework, Optuna \cite{akiba2019optuna}. Unlike traditional grid search, Optuna utilizes more efficient  Bayesian optimization  to navigate the hyperparameter space \cite{akiba2019optuna}. All hyperparameter search options and outcomes are stored in a database for a  comprehensive tuning process. The chosen hyperparameters are available in our \href{https://github.com/beckslopez/StudentMEH-Fitbit-ML}{GitHub} repository. 

\subsection{Evaluation Methodology and Metrics}

Due to the limited number of participants, we implement a {\it leave-one-participant-out cross-validation} strategy. In this approach, we exclude all data pertaining to one participant in each iteration wherein one participant may have several data instances. In each iteration, we utilize a participant's data as the test set while using the remaining participants'  data as the training set \cite{Wang2016, Sheikh2021, Adler2022}.  Note, prediction is at the weekly level.

We  calculate  the \(TP\)  true positives, \(FN\)  false negatives, \(TN\)  true negatives, and \(FP\) false positives for each test set in every iteration. These values are aggregated across all iterations to compute the final performance metrics at the end of cross-validation. To create confidence intervals, we utilize a bootstrap resampling method in which we keep the model fixed and resample the test set predictions \cite{rajkomar2018scalable}. 

Following standard practice, we utilize balanced accuracy (BA) and F1 to evaluate our classifiers. Sensitivity, also known as Recall, measures the proportion of TP correctly identified, while specificity measures the proportion of TN correctly identified. Meanwhile, balanced accuracy is their average.

\begin{equation}
   \text{Sensitivity} = \frac{TP}{TP + FN}, ~~ \text{Specificity} = \frac{TN}{TN + FP}
\end{equation}

\begin{equation}
    \text{Balanced Accuracy} = \frac{1}{2} \left( Sensitivity + Specificity\right)
\end{equation}

Precision, also known as positive predictive value, measures   TP  divided by  total number of predictions. F1 score is  the harmonic mean of precision and recall.

\begin{equation}
   \text{Precision} = \frac{TP}{TP + FP}, ~ F1 = 2 \times \frac{Prec \times Recall}{Prec + Recall}
\end{equation} 

For compactness, BA and F1 score are reported in the paper, while alternate metrics can be found in our \href{https://github.com/beckslopez/StudentMEH-Fitbit-ML}{GitHub} repository.

\begin{table*}[t]
  \centering
  \caption{Depression (CESD-10) classification  balanced accuracy and F1 scores.}
  \setlength{\tabcolsep}{3pt}
  \renewcommand{\arraystretch}{1.0} 
\begin{tabular}{@{}c|cccccc|cc@{}}
    \toprule
\textbf{BA} & \multicolumn{6}{c}{\underline{\textbf{Aggregation Level}}}\\
    \cline{1-1}\\ [-2ex]
   \textbf{Modality} & 1 & 4 & 6 & 8 & 12 & 24 & Best Agg. & Best Model \\
    \midrule
all modalities & 0.51 $\pm$ 0.02 & \textit{0.55 $\pm$ 0.04} & 0.54 $\pm$ 0.04 & 0.54 $\pm$ 0.05 & \textit{0.55 $\pm$ 0.04} & 0.47 $\pm$ 0.03 & 0.55 $\pm$ 0.04 & DT \\
calories & 0.56 $\pm$ 0.03 & 0.50 $\pm$ 0.05 & \textbf{0.62 $\pm$ 0.04} & \textbf{\textit{0.70 $\pm$ 0.04}} & 0.55 $\pm$ 0.04 & 0.47 $\pm$ 0.03 & 0.70 $\pm$ 0.04 & AdaBoost \\
distance & 0.49 $\pm$ 0.03 & 0.55 $\pm$ 0.04 & 0.48 $\pm$ 0.04 & \textit{0.58 $\pm$ 0.05} & 0.43 $\pm$ 0.03 & 0.52 $\pm$ 0.03 & 0.58 $\pm$ 0.05 & SVM \\
heart & 0.54 $\pm$ 0.05 & 0.58 $\pm$ 0.03 & 0.55 $\pm$ 0.03 & 0.57 $\pm$ 0.04 & \textit{\textbf{0.63 $\pm$ 0.04}} & 0.53 $\pm$ 0.04 & 0.63 $\pm$ 0.04 & AdaBoost \\
sleep & \textbf{0.63 $\pm$ 0.03} & \textbf{0.59 $\pm$ 0.04} & 0.61 $\pm$ 0.04 & \textit{0.69 $\pm$ 0.03} & 0.60 $\pm$ 0.05 & \textbf{0.67 $\pm$ 0.04} & 0.69 $\pm$ 0.03 & SVM \\
step & 0.56 $\pm$ 0.03 & 0.55 $\pm$ 0.04 & 0.58 $\pm$ 0.05 & \textit{0.69 $\pm$ 0.04} & 0.52 $\pm$ 0.04 & 0.56 $\pm$ 0.03 & 0.69 $\pm$ 0.04 & AdaBoost \\
\midrule
Best Modality & 0.63 $\pm$ 0.03 & 0.59 $\pm$ 0.04 & 0.62 $\pm$ 0.04 & 0.70 $\pm$ 0.04 & 0.63 $\pm$ 0.04 & 0.67 $\pm$ 0.04 & 0.70 $\pm$ 0.04 & AdaBoost \\
Best Model & RF & RF & SVM & AdaBoost & AdaBoost & SVM & AdaBoost \\
\midrule
       \midrule   
\textbf{F1} & \multicolumn{6}{c}{\underline{\textbf{Aggregation Level}}}\\
    \cline{1-1}\\ [-2ex]
   \textbf{Modality} & 1 & 4 & 6 & 8 & 12 & 24 & Best Agg. & Best Model \\
    \midrule
all modalities & 0.67 $\pm$ 0.05 & 0.67 $\pm$ 0.04 & \textit{0.69 $\pm$ 0.05} & 0.68 $\pm$ 0.04 & 0.65 $\pm$ 0.05 & 0.68 $\pm$ 0.05 & 0.69 $\pm$ 0.05 & AdaBoost \\
calories & 0.70 $\pm$ 0.05 & 0.69 $\pm$ 0.05 & 0.72 $\pm$ 0.03 & \textit{0.74 $\pm$ 0.04} & 0.66 $\pm$ 0.05 & 0.70 $\pm$ 0.04 & 0.74 $\pm$ 0.04 & AdaBoost \\
distance & 0.62 $\pm$ 0.05 & 0.72 $\pm$ 0.04 & 0.69 $\pm$ 0.05 & \textit{\textbf{0.78 $\pm$ 0.04}} & 0.69 $\pm$ 0.05 & 0.57 $\pm$ 0.05 & 0.78 $\pm$ 0.04 & SVM \\
heart & 0.65 $\pm$ 0.04 & 0.59 $\pm$ 0.04 & \textit{0.69 $\pm$ 0.04} & 0.57 $\pm$ 0.05 & 0.64 $\pm$ 0.05 & 0.68 $\pm$ 0.04 & 0.69 $\pm$ 0.04 & LR \\
sleep & \textbf{0.76 $\pm$ 0.03} & \textbf{0.78 $\pm$ 0.04} & \textbf{0.74 $\pm$ 0.04 }& 0.76 $\pm$ 0.04 & \textbf{0.76 $\pm$ 0.04} & \textbf{\textit{0.78 $\pm$ 0.03}} & 0.78 $\pm$ 0.03 & SVM \\
step & 0.71 $\pm$ 0.04 & 0.63 $\pm$ 0.05 & 0.69 $\pm$ 0.04 & \textit{0.77 $\pm$ 0.04} & 0.60 $\pm$ 0.04 & \textit{0.77 $\pm$ 0.04} & 0.77 $\pm$ 0.04 & AdaBoost \\
\midrule
Best Modality & 0.76 $\pm$ 0.03 & 0.78 $\pm$ 0.04 & 0.74 $\pm$ 0.04 & 0.78 $\pm$ 0.04 & 0.76 $\pm$ 0.04 & 0.78 $\pm$ 0.03 & 0.78 $\pm$ 0.03 & SVM \\
Best Model & RF & RF & XGBoost & SVM & LR & SVM & SVM \\
\bottomrule
\end{tabular}
\newline \footnotesize{ 
Bold indicates modality with the highest score for each column; italics indicate aggregation level with the highest score for each row. 
}
\label{table-depression}
\end{table*}

\begin{table*}[ht]
  \centering
  \setlength{\tabcolsep}{3pt}
  \caption{Anxiety (STAI) classification  Balanced Accuracy and F1 scores.}
  \label{table-anxiety}
  \renewcommand{\arraystretch}{1.0} 
  \begin{tabular}{@{}c|cccccc|c|c@{}}
    \toprule
\textbf{BA} & \multicolumn{6}{c}{\underline{\textbf{Aggregation Level}}}\\
    \cline{1-1}\\ [-2ex]
   \textbf{Modality} & 1 & 4 & 6 & 8 & 12 & 24 & Best Agg. & Best Model \\
    \midrule
all modalities & \textit{0.57 $\pm$ 0.04} & \textit{0.57 $\pm$ 0.04} & 0.52 $\pm$ 0.04 & 0.54 $\pm$ 0.04 & 0.53 $\pm$ 0.03 & 0.50 $\pm$ 0.04 & 0.57 $\pm$ 0.04 & AdaBoost \\
calories & \textit{0.58 $\pm$ 0.04} & 0.51 $\pm$ 0.04 & 0.57 $\pm$ 0.03 & 0.57 $\pm$ 0.04 & 0.50 $\pm$ 0.04 & 0.54 $\pm$ 0.04 & 0.58 $\pm$ 0.04 & LR \\
distance & \textbf{\textit{0.66 $\pm$ 0.0}}5 & 0.56 $\pm$ 0.03 & 0.56 $\pm$ 0.05 & 0.55 $\pm$ 0.04 & 0.56 $\pm$ 0.04 & \textbf{0.59 $\pm$ 0.04} & 0.66 $\pm$ 0.05 & RF \\
heart & 0.52 $\pm$ 0.04 & 0.53 $\pm$ 0.03 & \textbf{\textit{0.65 $\pm$ 0.03}} & 0.54 $\pm$ 0.05 & 0.59 $\pm$ 0.04 & 0.54 $\pm$ 0.02 & 0.65 $\pm$ 0.03 & DT \\
sleep & 0.63 $\pm$ 0.04 & \textbf{\textit{0.64 $\pm$ 0.03}} & 0.60 $\pm$ 0.05 & \textbf{0.60 $\pm$ 0.04} & 0.54 $\pm$ 0.04 & 0.58 $\pm$ 0.04 & 0.64 $\pm$ 0.03 & XGBoost \\
step & 0.61 $\pm$ 0.05 & \textit{0.63 $\pm$ 0.04} & 0.61 $\pm$ 0.04 & 0.60 $\pm$ 0.05 & \textbf{0.62 $\pm$ 0.03} & 0.57 $\pm$ 0.04 & 0.63 $\pm$ 0.04 & AdaBoost \\
\midrule
Best Modality & 0.66 $\pm$ 0.05 & 0.64 $\pm$ 0.03 & 0.65 $\pm$ 0.03 & 0.60 $\pm$ 0.04 & 0.62 $\pm$ 0.03 & 0.59 $\pm$ 0.04 & 0.66 $\pm$ 0.05 & RF \\
\midrule
Best Model & RF & XGBoost & DT & AdaBoost & LR & SVM & RF \\
\midrule
    \midrule   
\textbf{F1} & \multicolumn{6}{c}{\underline{\textbf{Aggregation Level}}}\\
    \cline{1-1}\\ [-2ex]
   \textbf{Modality} & 1 & 4 & 6 & 8 & 12 & 24 & Best Agg. & Best Model \\
    \midrule
all modalities & 0.72 $\pm$ 0.04 & 0.72 $\pm$ 0.04 & 0.63 $\pm$ 0.05 & 0.66 $\pm$ 0.04 & 0.74 $\pm$ 0.05 & \textit{0.77 $\pm$ 0.04} & 0.77 $\pm$ 0.04 & AdaBoost \\
calories & \textit{0.75 $\pm$ 0.04} & 0.73 $\pm$ 0.04 & 0.73 $\pm$ 0.04 & 0.62 $\pm$ 0.04 & 0.68 $\pm$ 0.05 & 0.72 $\pm$ 0.05 & 0.75 $\pm$ 0.04 & LR \\
distance & \textit{0.75 $\pm$ 0.04} & 0.72 $\pm$ 0.04 & 0.70 $\pm$ 0.04 & 0.74 $\pm$ 0.03 & 0.74 $\pm$ 0.04 & 0.72 $\pm$ 0.04 & 0.75 $\pm$ 0.04 & RF \\
heart & 0.68 $\pm$ 0.04 & 0.76 $\pm$ 0.04 & 0.78 $\pm$ 0.03 & 0.60 $\pm$ 0.04 & 0.74 $\pm$ 0.04 & \textbf{\textit{0.79 $\pm$ 0.04}} & 0.79 $\pm$ 0.04 & AdaBoost \\
sleep & \textbf{\textit{0.79 $\pm$ 0.03}} & \textbf{0.78 $\pm$ 0.04} & 0.74 $\pm$ 0.03 & 0.77 $\pm$ 0.03 & 0.72 $\pm$ 0.04 & 0.67 $\pm$ 0.04 & 0.79 $\pm$ 0.03 & RF \\
step & 0.75 $\pm$ 0.03 & 0.75 $\pm$ 0.03 & \textbf{\textit{0.83 $\pm$ 0.03}} & \textbf{0.78 $\pm$ 0.03} & \textbf{0.77 $\pm$ 0.04} & 0.70 $\pm$ 0.04 & 0.83 $\pm$ 0.03 & RF \\
\midrule
Best Modality & 0.79 $\pm$ 0.03 & 0.78 $\pm$ 0.04 & 0.83 $\pm$ 0.03 & 0.78 $\pm$ 0.03 & 0.77 $\pm$ 0.04 & 0.79 $\pm$ 0.04 & 0.83 $\pm$ 0.03 & RF \\
\midrule
Best Model & RF & XGBoost & RF & SVM & LR & AdaBoost & RF \\
\bottomrule
\end{tabular}
\newline \footnotesize{ 
Bold indicates modality with the highest score for each column; italics indicate aggregation level with the highest score for each row.} 
\end{table*}

\begin{table*}[ht]
  \centering
  \setlength{\tabcolsep}{3pt}
  \caption{Stress (PSS-4) classification  balanced accuracy and F1 scores.}
  \label{table-stress}
  \renewcommand{\arraystretch}{1.0} 
  \begin{tabular}{@{}c|cccccc|c|c@{}}
        \toprule
\textbf{BA} & \multicolumn{6}{c}{\underline{\textbf{Aggregation Level}}}\\
    \cline{1-1}\\ [-2ex]
   \textbf{Modality} & 1 & 4 & 6 & 8 & 12 & 24 & Best Agg. & Best Model \\
    \midrule
all modalities & 0.55 $\pm$ 0.04 & 0.52 $\pm$ 0.04 & 0.52 $\pm$ 0.04 & \textit{0.56 $\pm$ 0.03} & 0.50 $\pm$ 0.04 & 0.55 $\pm$ 0.05 & 0.56 $\pm$ 0.03 & DT \\
calories & \textit{\textbf{0.73 $\pm$ 0.04}} & 0.61 $\pm$ 0.04 & 0.63 $\pm$ 0.04 & 0.56 $\pm$ 0.04 & 0.58 $\pm$ 0.06 & 0.57 $\pm$ 0.03 & 0.73 $\pm$ 0.04 & AdaBoost \\
distance & 0.50 $\pm$ 0.04 & 0.52 $\pm$ 0.05 & 0.55 $\pm$ 0.04 & 0.52 $\pm$ 0.03 & 0.56 $\pm$ 0.04 & \textit{0.60 $\pm$ 0.04} & 0.60 $\pm$ 0.04 & RF \\
heart & 0.57 $\pm$ 0.04 & 0.57 $\pm$ 0.05 & \textbf{0.67 $\pm$ 0.03} & \textbf{0.65 $\pm$ 0.04} & \textbf{\textit{0.69 $\pm$ 0.03}} & 0.59 $\pm$ 0.04 & 0.69 $\pm$ 0.03 & AdaBoost \\
sleep & 0.64 $\pm$ 0.05 & 0.64 $\pm$ 0.03 & \textit{0.66 $\pm$ 0.03} & 0.59 $\pm$ 0.04 & 0.63 $\pm$ 0.04 & \textbf{0.62 $\pm$ 0.03} & 0.66 $\pm$ 0.03 & SVM \\
step & 0.52 $\pm$ 0.04 & \textit{\textbf{0.65 $\pm$ 0.04}} & 0.54 $\pm$ 0.04 & 0.47 $\pm$ 0.04 & 0.57 $\pm$ 0.04 & 0.59 $\pm$ 0.04 & 0.65 $\pm$ 0.04 & DT \\
\midrule
Best Modality & 0.73 $\pm$ 0.04 & 0.65 $\pm$ 0.04 & 0.67 $\pm$ 0.03 & 0.65 $\pm$ 0.04 & 0.69 $\pm$ 0.03 & 0.62 $\pm$ 0.03 & 0.73 $\pm$ 0.04 & AdaBoost \\
\midrule
Best Model & AdaBoost & DT & SVM & LR & AdaBoost & RF & AdaBoost \\
\midrule
\midrule
\textbf{F1} & \multicolumn{6}{c}{\underline{\textbf{Aggregation Level}}}\\
    \cline{1-1}\\ [-2ex]
   \textbf{Modality} & 1 & 4 & 6 & 8 & 12 & 24 & Best Agg. & Best Model \\
    \midrule
all modalities & 0.65 $\pm$ 0.05 & 0.61 $\pm$ 0.05 & 0.63 $\pm$ 0.05 & 0.61 $\pm$ 0.05 & 0.62 $\pm$ 0.05 & \textit{0.68 $\pm$ 0.05} & 0.68 $\pm$ 0.05 & AdaBoost \\
calories & 0.74 $\pm$ 0.04 & \textbf{\textit{0.81 $\pm$ 0.04}} & 0.66 $\pm$ 0.05 & 0.71 $\pm$ 0.04 & 0.75 $\pm$ 0.03 & 0.59 $\pm$ 0.05 & 0.81 $\pm$ 0.04 & AdaBoost \\
distance & \textit{0.67 $\pm$ 0.05} & 0.66 $\pm$ 0.04 & 0.57 $\pm$ 0.05 & 0.57 $\pm$ 0.05 & 0.66 $\pm$ 0.04 & 0.66 $\pm$ 0.04 & 0.67 $\pm$ 0.05 & AdaBoost \\
heart & 0.70 $\pm$ 0.04 & 0.71 $\pm$ 0.05 & \textbf{0.74 $\pm$ 0.04} & 0.72 $\pm$ 0.04 & \textbf{\textit{0.77 $\pm$ 0.03}} & \textbf{0.72 $\pm$ 0.04} & 0.77 $\pm$ 0.03 & AdaBoost \\
sleep & \textbf{0.76 $\pm$ 0.04} & \textit{0.80 $\pm$ 0.03} & 0.71 $\pm$ 0.04 & \textbf{0.73 $\pm$ 0.04} & 0.71 $\pm$ 0.05 & 0.70 $\pm$ 0.04 & 0.80 $\pm$ 0.03 & AdaBoost \\
step & 0.66 $\pm$ 0.04 & \textit{0.77 $\pm$ 0.03} & 0.68 $\pm$ 0.05 & 0.64 $\pm$ 0.04 & 0.65 $\pm$ 0.05 & 0.59 $\pm$ 0.06 & 0.77 $\pm$ 0.03 & DT \\
\midrule
Best Modality & 0.76 $\pm$ 0.04 & 0.81 $\pm$ 0.04 & 0.74 $\pm$ 0.04 & 0.73 $\pm$ 0.04 & 0.77 $\pm$ 0.03 & 0.72 $\pm$ 0.04 & 0.81 $\pm$ 0.04 & AdaBoost \\
\midrule
Best Model & AdaBoost & AdaBoost & SVM & RF & AdaBoost & AdaBoost & AdaBoost \\
\bottomrule
\end{tabular}
\newline 
\footnotesize{Bold indicates modality with the highest score for each column; italics indicate aggregation level with the highest score for each row.} 
\end{table*}

\section{Results}
\begin{figure*}[htbp] 
    \centering

    \begin{minipage}[t]{0.48\textwidth}
        \centering
        \caption*{\textbf{(A) Performance Across Modalities}}
        \includegraphics[width=\linewidth]{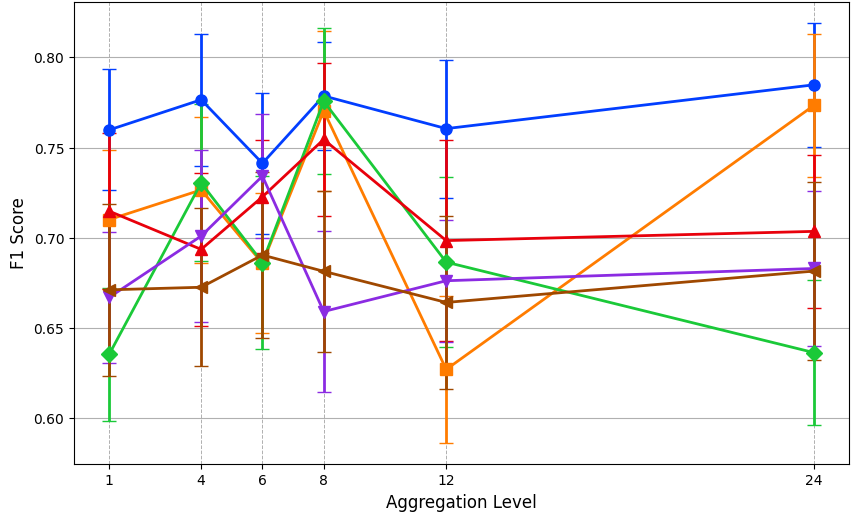}\vspace{0.3em}
        \includegraphics[width=\linewidth]{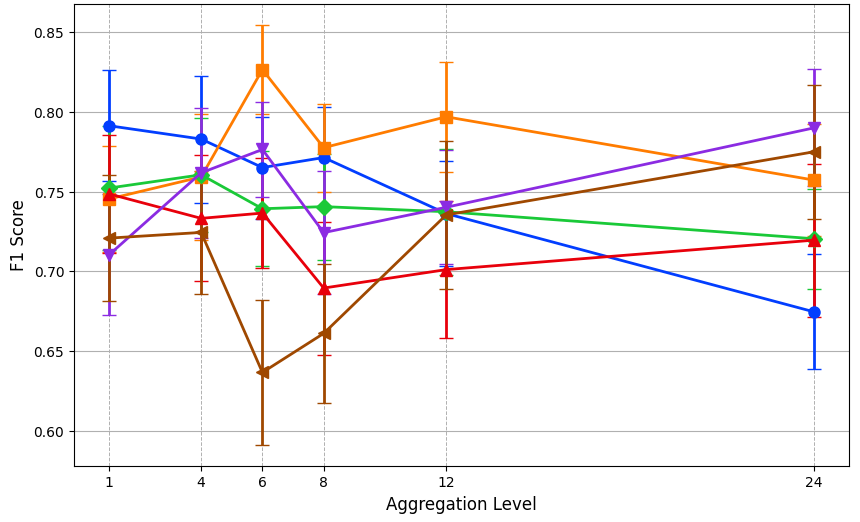}\vspace{0.3em}
        \includegraphics[width=\linewidth]{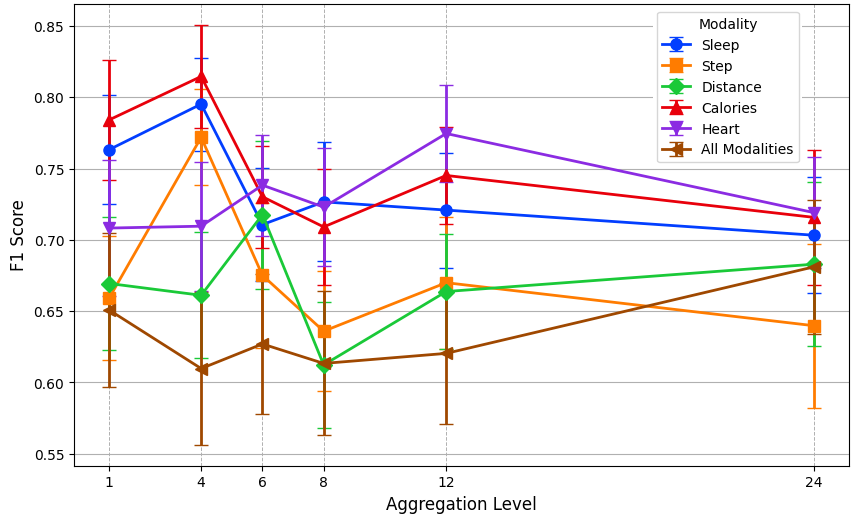}
    \end{minipage}
    \hfill
    \begin{minipage}[t]{0.48\textwidth}
        \centering
        \caption*{\textbf{(B) Feature Importance}}
        \includegraphics[width=\linewidth]{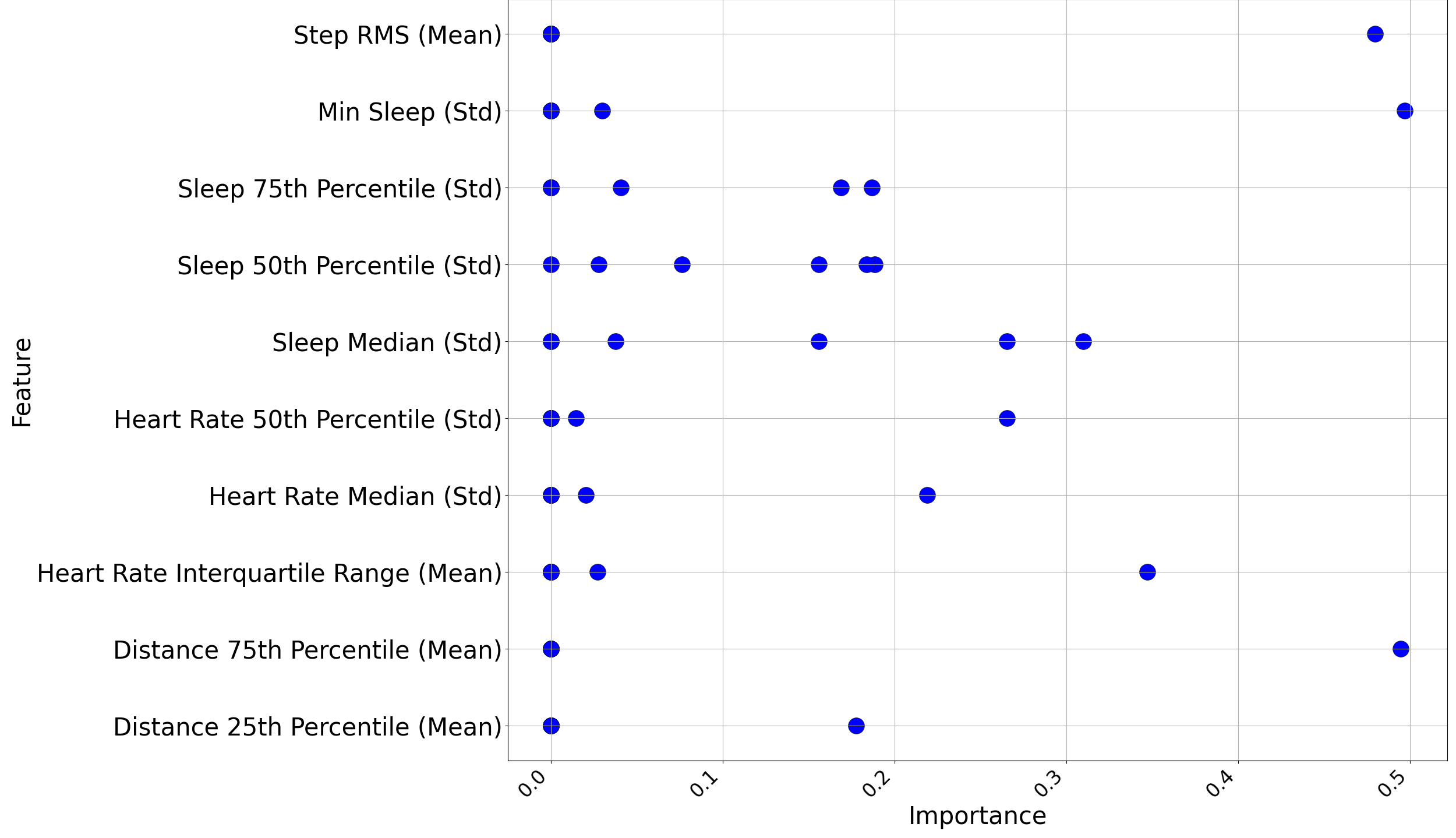}\vspace{0.3em}
        \includegraphics[width=\linewidth]{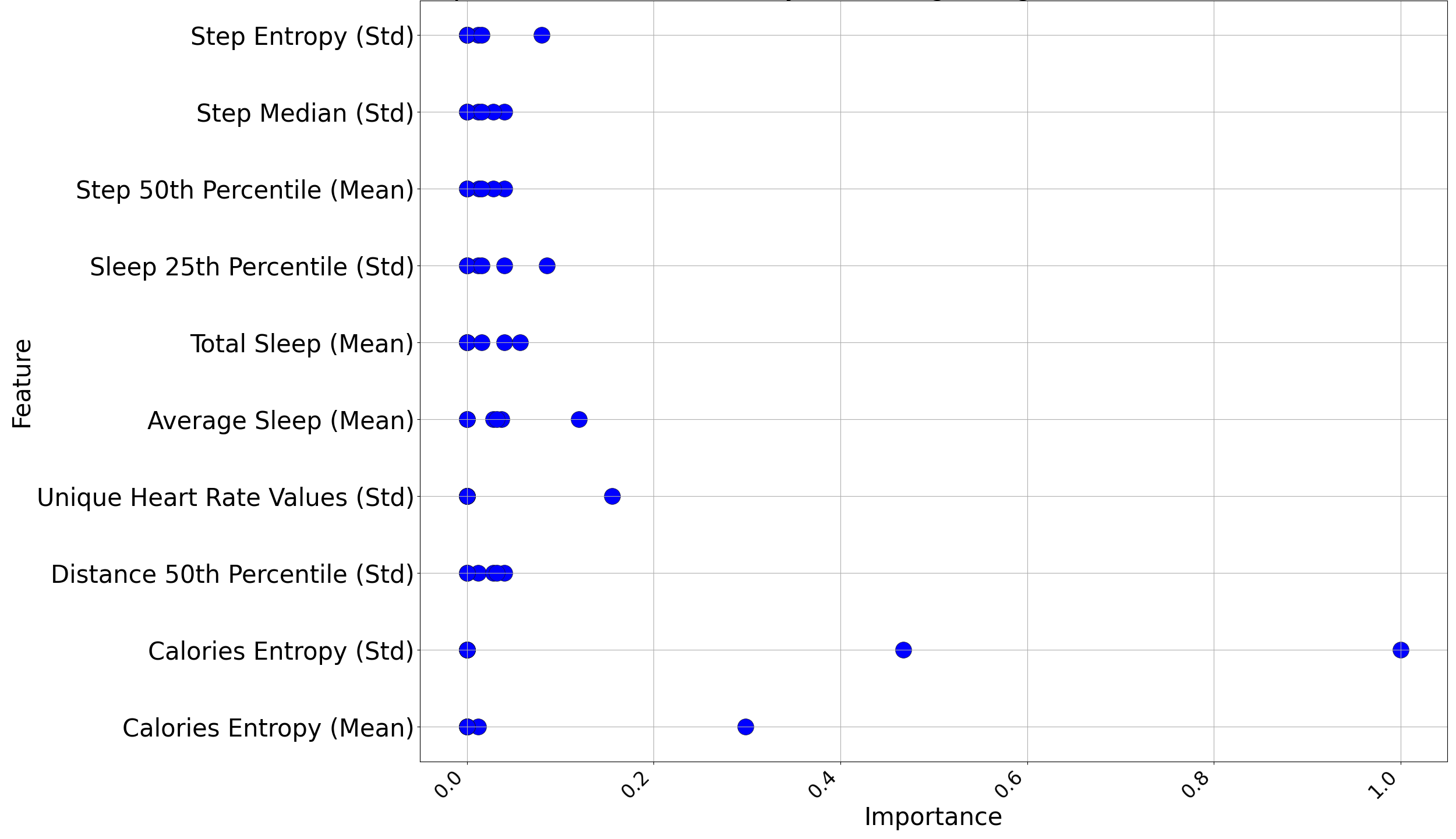}\vspace{0.3em}
        \includegraphics[width=\linewidth]{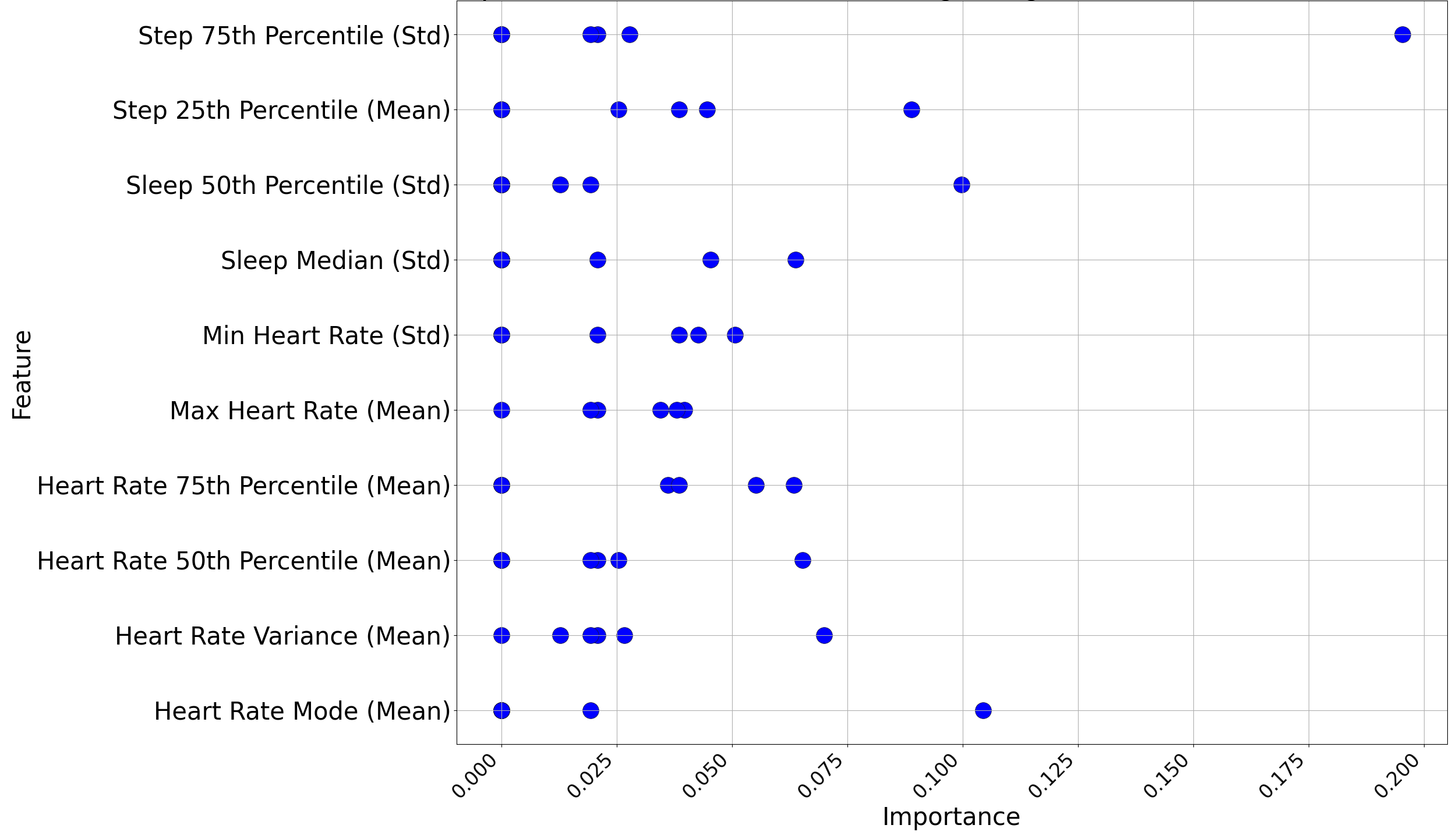}
    \end{minipage}

    \caption{(A) Performance across aggregation levels; (B) Feature importance for AdaBoost classifiers.}
    \label{fig:joined}
\end{figure*}

The performance of our classifiers for depression, anxiety, and stress screening using different data modalities and time level aggregations is detailed in Tables \ref{table-depression}, \ref{table-anxiety}, and \ref{table-stress}, respectively. Modalities include calories, distance, heart, sleep, step, and a combination of all modalities. The performance of these models is measured at time intervals of 1, 4, 6, 8, 12, and 24 hours - covering slices of daily living (morning versus afternoon, night versus day, etc.). Line plots of F1 Scores from Tables \ref{table-depression}, \ref{table-anxiety}, \ref{table-stress}, highlighting visually trends across aggregation levels, are in Figure \ref{fig:joined} and our \href{https://github.com/beckslopez/StudentMEH-Fitbit-ML}{GitHub} repository. 

\subsection{Depression, Anxiety, and Stress Classification}
For depression screening with BA (Table \ref{table-depression}), AdaBoost emerged as the top classifier across different modalities and time intervals. The highest BA of $0.70 \pm 0.04$ was achieved by AdaBoost with 8-hour aggregation of calories, though sleep and step were almost as successful. These high BA scores suggest that capturing physical activity at the 8-hour granularity is good for depression screening. 

In contrast, higher F1 scores of $0.78$ were consistently achieved for all time aggregation levels and multiple classifiers, indicating the robustness of the performance. Similarly, most individual modalities, 
excluding heart rate, demonstrate strong performance - with distance and sleep posting the highest F1. Note that we achieve the $0.74$ to $0.78$ range for sleep across all time aggregations, demonstrating its high potential for classifying depression. Overall, step count and sleep are 
promising modalities for depression screening.

For anxiety classification (Table \ref{table-anxiety}), the BA metric revealed that Random Forest (RF) was the top performing classifier. The highest BA of $0.66 \pm 0.05$ was achieved by RF at the 1 hour interval for the distance modality, though 4 and 6 hour time intervals were also strong performers for heart rate and sleep, respectively. This suggests that RF is particularly effective at shorter observation periods, such as the 1-hour to 6-hour interval, for achieving strong BA performance. The highest  F1 score of $0.83 \pm 0.03$ was achieved at the 6-hour interval for step by RF. Though all individual modalities hold a strong signal for anxiety screening in 
the 0.75 to 0.83 F1 range. 

For stress classification (Table \ref{table-stress}), AdaBoost applied to the  calories spent aggregated at the 1-hour time
interval emerged as the top performer with a BA of $0.73$, a higher BA than when screening than for depression or anxiety. AdaBoost also came in as second best performer, achieving a BA of $0.69 \pm 0.03$ at the 12-hour interval for the heart rate modality. 

We also observe the strong F1 performance for stress classification at $0.81 \pm 0.04$  for the calories modality at a 4 hour interval followed closely by $0.77\pm0.03$ for heart at a 12 hour interval. Overall, variations in the time granularity for aggregation did not have a strong  impact, achieving scores above 0.72 F1 or higher for all time intervals.  

The concatenation of all modalities led to a suboptimal performance compared to the top individual modalities despite using the same model optimization strategies. This could be due to conflicting signals between modalities found in the considerably higher dimensionality of the data when processed in this manner. In the context of our study, we thus recommend to carefully review and select a single modality such as heart or sleep over indiscriminately working
with all of them. 

\begin{table}[t]
\centering
\caption{Number of times each model was best for each aggregation levels for each condition and modality combination. A value of 6 indicates the model won across all aggregation levels. 
}
\setlength{\tabcolsep}{3pt}
\begin{tabular}{|c|c|c|c|c|c|c|c|c|}
\hline
Condition & Model & All & Cal. & Dist. & Heart & Sleep & Step & \textbf{Sum} \\
\hline
\multirow{6}{*}{Depression} & AdaBoost & 5 & 4 & 2 & 1 & 0 & 4 & \textbf{16} \\
\cline{2-9}
& DT          & 1 & 0 & 0 & 0 & 0 & 0 &\textbf{ 1} \\
\cline{2-9}
& SVM         & 0 & 2 & 4 & 2 & 1 & 2 &\textbf{ 11} \\
\cline{2-9}
& LR          & 0 & 0 & 0 & 1 & 2 & 0 & \textbf{3 }\\
\cline{2-9}
& XGBoost     & 0 & 0 & 0 & 1 & 1 & 0 & \textbf{2} \\
\cline{2-9}
& RF          & 0 & 0 & 0 & 1 & 2 & 0 & \textbf{3} \\
\hhline{|=========|}
\multirow{6}{*}{Stress} & AdaBoost & 1 & 3 & 3 & 2 & 2 & 0 & \textbf{11} \\
\cline{2-9}
& DT          & 3 & 0 & 1 & 0 & 0 & 2 & \textbf{6} \\
\cline{2-9}
& SVM         & 0 & 3 & 1 & 1 & 1 & 2 & \textbf{8} \\
\cline{2-9}
& LR          & 1 & 0 & 1 & 3 & 0 & 0 & \textbf{5} \\
\cline{2-9}
& XGBoost     & 1 & 0 & 0 & 0 & 0 & 0 & \textbf{1} \\
\cline{2-9}
& RF          & 0 & 0 & 0 & 0 & 3 & 2 & \textbf{5} \\ \hhline{|=========|}
\multirow{6}{*}{Anxiety} & AdaBoost & 6 & 3 & 2 & 1 & 2 & 1 & \textbf{15} \\
\cline{2-9}
& DT          & 0 & 0 & 0 & 1 & 0 & 0 & \textbf{1} \\
\cline{2-9}
& SVM         & 0 & 0 & 0 & 2 & 0 & 2 & \textbf{4} \\
\cline{2-9}
& LR          & 0 & 3 & 2 & 0 & 1 & 1 & \textbf{7} \\
\cline{2-9}
& XGBoost     & 0 & 0 & 0 & 2 & 2 & 0 & \textbf{4} \\
\cline{2-9}
& RF          & 0 & 0 & 2 & 0 & 1 & 2 & \textbf{5} \\
\hline
\end{tabular}
\label{table:winning_model}
\end{table}

\begin{table*}[ht]
\centering
\setlength{\tabcolsep}{4pt}
\caption{Highest F1 Score and corresponding balanced accuracy (BA) achieved by each model across aggregation levels.}
\begin{tabular}{|c|c|c|c|c|c|c|c|c|c|c|c|c|c|}
\hline
\multirow{2}{*}{Condition} & \multirow{2}{*}{Model} & \multicolumn{2}{|c|}{All Modalities} & \multicolumn{2}{|c|}{Calories} & \multicolumn{2}{|c|}{Distance} & \multicolumn{2}{|c|}{Heart} & \multicolumn{2}{|c|}{Sleep} & \multicolumn{2}{|c|}{Step} \\
\cline{3-14}
& & F1 & BA & F1 & BA & F1 & BA & F1 & BA & F1 & BA & F1 & BA \\
\hline

\multirow{6}{*}{Depression} 
& AdaBoost         & \textbf{0.69} & 0.54 & \textbf{0.75} & \textbf{0.71} & 0.73 & 0.49 & 0.70 & 0.48 & 0.74 & 0.55 & \textbf{0.77} & \textbf{0.69} \\ \cline{2-14}
& DecisionTree     & 0.67 & \textbf{0.55} & 0.64 & 0.56 & 0.66 & 0.48 & 0.62 & 0.53 & 0.72 & 0.57 & 0.67 & 0.55 \\ \cline{2-14}
& SVM              & 0.59 & 0.46 & 0.72 & 0.62 & \textbf{0.78} & \textbf{0.58} & \textbf{0.73} & 0.53 & \textbf{0.78} & \textbf{0.67} & 0.71 & 0.58 \\ \cline{2-14}
& LogisticReg      & 0.59 & 0.46 & 0.71 & 0.61 & 0.62 & 0.48 & 0.66 & 0.54 & \textbf{0.78} & 0.63 & 0.67 & 0.52 \\ \cline{2-14}
& XGBoost          & 0.61 & 0.47 & 0.61 & 0.45 & 0.71 & \textbf{0.58 }& 0.68 & \textbf{0.57} & 0.74 & 0.61 & 0.63 & 0.55 \\ \cline{2-14}
& RandomForest     & 0.62 & 0.50 & 0.62 & 0.46 & 0.67 & 0.49 & 0.68 & 0.53 & 0.78 & 0.63 & 0.60 & 0.52 \\ \hhline{|==============|}

\multirow{6}{*}{Stress} 
& AdaBoost         & \textbf{0.68} & \textbf{0.55} & 0.81 & \textbf{0.75} & 0.72 & \textbf{0.71} & \textbf{0.77} & \textbf{0.69} & \textbf{0.80} & 0.64 & 0.67 & 0.57 \\ \cline{2-14}
& DecisionTree     & 0.65 & \textbf{ 0.55} & 0.63 & 0.51 & 0.66 & 0.52 & 0.66 & 0.53 & 0.77 & \textbf{0.65} & \textbf{0.77} & \textbf{0.65} \\ \cline{2-14}
& SVM              & 0.64 & 0.49 & \textbf{ 0.75} & 0.58 & 0.64 & 0.52 & 0.74 & 0.67 & 0.66 & 0.52 & 0.67 & 0.53 \\ \cline{2-14}
& LogisticReg      & 0.62 & 0.50 & 0.70 & 0.56 & 0.70 & 0.56 & 0.72 & 0.65 & 0.69 & 0.57 & 0.65 & 0.41 \\ \cline{2-14}
& XGBoost          & 0.61 & 0.52 & 0.63 & 0.46 & 0.61 & 0.52 & 0.70 & 0.57 & 0.71 & 0.57 & 0.65 & 0.51 \\ \cline{2-14}
& RandomForest     & 0.63 & 0.49 & 0.61 & 0.50 & \textbf{0.77} & 0.65 & 0.71 & 0.59 & 0.77 & \textbf{0.65} & 0.68 & 0.54 \\ \hhline{|==============|}

\multirow{6}{*}{Anxiety} 
& AdaBoost         & \textbf{0.77} & 0.57 & 0.76 & 0.54 & \textbf{0.76} & 0.54 & \textbf{0.79} & 0.54 & 0.77 & 0.60 & 0.75 & \textbf{0.61} \\ \cline{2-14}
& DecisionTree     & 0.69 & 0.55 & 0.65 & 0.52 & 0.72 & 0.52 & 0.78 & \textbf{0.65} & 0.72 & 0.56 & 0.74 & 0.58 \\ \cline{2-14}
& SVM              & 0.61 & 0.47 & \textbf{0.78} & \textbf{0.60} & 0.72 & \textbf{0.59} & 0.76 & 0.53 & 0.78 & 0.60 & 0.78 & 0.60 \\ \cline{2-14}
& LogisticReg      & 0.74 & \textbf{0.58} & 0.75 & 0.57 & 0.75 & 0.57 & 0.70 & 0.56 & 0.74 & 0.60 & 0.80 & 0.55 \\ \cline{2-14}
& XGBoost          & 0.70 & 0.53 & 0.70 & 0.50 & 0.74 & 0.55 & 0.76 & 0.54 & 0.78 & 0.64 & 0.72 & 0.59 \\ \cline{2-14}
& RandomForest     & 0.69 & 0.49 & 0.64 & 0.43 & \textbf{0.76} & 0.54 & 0.73 & 0.49 & \textbf{0.79} & \textbf{0.65} & \textbf{0.83} & \textbf{0.61} \\ \hline

\end{tabular}
\label{table:winning_model_scores}
\newline \footnotesize{Bold indicates the model with the highest score for each modality and condition.}
\end{table*}

We further focus on the detailed results of model performance across aggregation levels for each modality-label combination, as displayed in Table \ref{table:winning_model}. Here, we include the counts of the number of times each model had the highest F1 per each aggregation level for each condition and modality combination. In addition, the best of the F1 scores and its matching BA scores is then included in Table \ref{table:winning_model_scores}.  

For depression, AdaBoost excels particularly in the calories and step modalities, winning 4 times in each category. This indicates its stability and effectiveness in these contexts. SVM shows strong performance with distance data, where it wins 4 times.
This suggests it is well-suited for capturing patterns in this modality. However, the heart modality shows more evenly
distributed
performance, with SVM, Logistic Regression, XGBoost, and Random Forest each achieving 1 win.
This hints at the complexity of heart-related data that may benefit from diverse modeling approaches. In the sleep modality, no single model dominates, with Random Forest leading with 2 wins, and Logistic Regression, SVM, and XGBoost each securing 1 win.
This indicates a varied set of models can perform well.

For anxiety, AdaBoost continues to be a strong performer, particularly in calories and distance, winning 3 and 2 times, respectively. This indicates the robustness of AdaBoost in these contexts. However, the heart modality shows a more balanced competition, with XGBoost and SVM leading, each winning 2 times, while Decision Tree and AdaBoost each achieve 1 win. In the step modality, Random Forest secures 2 wins, while SVM, AdaBoost, and Logistic Regression each have 1 win. This suggests that step-based patterns for anxiety detection may require multiple modeling approaches. 

For stress, the results are more varied across modalities. Decision Tree stands out with 3 wins in the distance category, showing strength in capturing patterns from this data type. AdaBoost and SVM both perform well in the calories modality, each winning 3 times.
This highlights their ability to handle stress-related signals effectively. With three wins each, logistic regression and random forest models are best for heart rate and sleep modalities, respectively. The step modality had three models tie with 2 wins: Decision Tree, SVM, and Random Forest. This suggests the need for varied modeling strategies to capture stress-related patterns.

These results highlight that AdaBoost and SVM are the best models. AdaBoost is a consistently strong performer across various modalities, particularly for depression and anxiety screening, indicating its robustness and versatility. Meanwhile, SVM is more competitive in specific contexts such as distance and step data for depression and stress, 
suggesting it can effectively capture subtle variations in these measurements.

\subsection{Feature Analysis}

Identifying the top features across all modalities is crucial for both interpretation and explainability, as it allows us to understand which factors consistently influence predictions in our model. To compute the top 10 features utilized by our
models, 
we used the AdaBoost model as it performed best across all modalities and conditions. 
For each feature, we obtained importance values as determined by the model across six aggregation level within each modality and condition combination. From there, we retained only those that ranked highest in importance at any one particular aggregation level.

As depicted in 
Figure \ref{fig:joined}
this feature importance analysis reveals interesting insights into the predictors of these mental health conditions. For depression, sleep and heart rate metrics dominate, highlighting the role of sleep consistency and heart rate stability in predicting depressive symptoms. Step and sleep metrics are particularly influential in anxiety prediction, with key features such as step entropy, median, and sleep percentiles indicating the importance of physical activity and sleep stability. Stress prediction relies heavily on heart rate features, suggesting a strong link between stress and heart rate variability. Across all three models, step, sleep, and heart rate data emerge as key predictors of mental health outcomes.

\section{Discussion}
\subsection{Ethical and Implementation Considerations}

The integration of Fitbit devices to monitor mental health introduces both ethical and implementation considerations. Young adults, especially those navigating the challenges of university life, often encounter stress from academic pressure, social dynamics, and the newfound independence that comes with young adulthood \cite{cage2021student}. Universities have a responsibility to support students’ mental well-being, ensuring that they are in a healthy state to pursue academic and personal goals. By using Fitbit data, institutions could gain valuable insights into stress levels, sleep patterns, physical activity, and other mental health indicators, enabling better allocation of resources or identify patterns that may signal a need for intervention. 

However, ethical concerns arise regarding student autonomy and privacy. Monitoring physiological and mental health data may feel intrusive to students \cite{rooksby2019student}, who might worry about how this data will be used. Consent is a key factor; students should have the clear choice to opt-in or opt-out of monitoring, with a full understanding of the data collected, the intended use, and who will access it. Overall, transparency about data collection, storage, and usage policies will be crucial for building trust between students and the universities. 

From an implementation perspective, the widespread use of Fitbits and other wearable devices offers an advantage, as many students are already familiar with these tools and may value their convenience. However, in an academic setting, using Fitbits for mental health monitoring changes the device’s role—from a personal wellness tool to a potential means of institutional oversight. To ensure that this monitoring remains supportive rather than intrusive, universities must establish clear limits on data usage, with privacy policies that prioritize students' control over their personal health data. 

\subsection{Limitations}
Fitbit devices are subject to user compliance and device functionality issues, requiring consistent wearing and charging - factors  prone to vary among participants. Non-adherence 
can result in gaps in data collection,  leading to data challenges. Additionally, Fitbit data is inherently limited by the capabilities of the sensors. For example, Fitbit’s sleep detection relies on accelerometer data and heart rate variability, which may not accurately reflect sleep patterns for every individual, especially in cases of atypical sleep behaviors or medical conditions \cite{haghayegh2019accuracy}. Sensor performance can vary between individuals due to factors such as body composition and movement patterns \cite{Feehan2018}.

Another limitation of this research is the lack of participants with continuous Fitbit data in the StudentMEH dataset, which is a frequent challenge in related research \cite{wang2014studentlife,zhang2021relationship,wang2018tracking,boukhechba2018demonicsalmon,tlachac2024symptom}. To overcome this limitation, we divided the data into weekly intervals, thus reducing spareness and increasing the overall data size. While this approach limits longitudinal analysis, it enables us to concentrate on identifying the most promising aggregation levels, modalities, and models. 

\subsection{Future Work Modeling Fitbit Data}
Given the robust nature of the StudentMEH Fitbit dataset, there are several future research avenues that we intend to pursue in order to enhance the scope and impact of our work. For example, we plan to explore intra-personal modeling where patterns are analyzed within each individual over time. This could reveal valuable insights into trends that may indicate an impending mental health crisis as well as student physiological and mental health responses to a pandemic. 

Our future work also includes leveraging more advanced models, such as deep neural networks. For instance, the small dataset limitation can be addressed using self-supervised learning, while higher predictive performance may be achieved using a multimodal training scheme. By treating the Fitbit data as a time series rather than extracted features, we could further use specialized methods designed to capture sequential patterns and trends over time. Time series models may  enable us to gain an understanding of how health indicators evolve over time as students progress through the academic year, approaching examinations 
and other important milestones. 

Our StudentMEH Fitbit dataset shares modalities with other student datasets, like the publicly available GLOBEM dataset \cite{xu2022globem}. Thus, our future work also includes determining the generalizability of modeling results across datasets, providing important information for widespread implementation. 

Lastly, we intend to expand upon our research by incorporating the environmental data in StudentMEH alongside Fitbit data to gain a more comprehensive view of factors influencing mental health. Environmental data, such as room conditions, air quality, noise levels, and so on, could provide context to individual behavioral patterns captured by the Fitbit. Given the spareness of the environmental data, a prerequisite for such research is understanding the most effective Fitbit modalities, time aggregations, and models for depression, anxiety, and stress screening - which has been our focus thus far. 

\section{Conclusion}
For this research, we prepared and analyzed data from the unique StudentMEH Fitbit dataset, which combines physiological data with mental health screening scores from 160 college students. An F1 of 0.78 was achieved with the sleep modality using an SVM model at a 4-hour aggregation level for depression screening, an F1 of 0.83 was achieved with the step modality using a Random Forest model at a 6-hour aggregation level for anxiety screening, and an F1 of 0.81 was achieved with the calories modality using AdaBoost at a 4-hour interval for stress screening. These results indicate the importance of sleep data in detecting depressive symptoms, physical activity patterns in detecting anxiety symptoms. Tree ensembles were overall the most effective models, consistently outperforming other classifiers across various conditions. Our research provides important modeling insights that promise to guide future research that uses physiological data to passively assess student mental health. 

\noindent \textit{Code Availability Statement:} All code along with our experimental settings will be made available in a \href{https://github.com/beckslopez/StudentMEH-Fitbit-ML}{GitHub} repository. 

\section*{Acknowledgments}
This research was supported by NSF \#2028224, NSF  \#10374216, NSF \#2349370, NSF  \#1852498, NIH NIMH \#P50MH129701, Henry Luce Foundation, and WPI Data Science Department.

\bibliographystyle{tran2}
\bibliography{references}
\end{document}